\documentclass[runningheads]{llncs}
\usepackage{graphicx}
\usepackage{amsmath,amssymb} % define this before the line numbering.
\usepackage{color}
\usepackage{dsfont}
\usepackage{commath}
\usepackage{multirow}
\usepackage{booktabs}
\usepackage{floatrow}
\usepackage{tabularx}
\usepackage{caption}
\usepackage{hhline}
\usepackage{subcaption}
\newcommand{\etal}{\textit{et al}.}

\newcolumntype{L}[1]{>{\raggedright\arraybackslash}p{#1}}
\newcolumntype{C}[1]{>{\centering\arraybackslash}p{#1}}
\newcolumntype{R}[1]{>{\raggedleft\arraybackslash}p{#1}}

\begin{document}
%===========================================================

\title{Cross-Resolution Person Re-identification with Deep Antithetical Learning}
\titlerunning{Cross-Resolution Person ReID}

%===========================================================

\author{Zijie Zhuang \and
Haizhou Ai \and
Long Chen \and
Chong Shang}

\authorrunning{Zijie Zhuang et al.} % A shorter version of authors' name

\institute{Tsinghua National Lab for Info. Sci. \& Tech. (TNList), \\
Department of Computer Science and Technology, Tsinghua University, Beijing, P.R.China, 100084 \\
\email{jayzhuang42@gmail.com}}
\maketitle

%===========================================================
\begin{abstract}
Images with different resolutions are ubiquitous in public person re-identification (ReID) datasets and real-world scenes, it is thus crucial for a person ReID model to handle the image resolution variations for improving its generalization ability.
However, most existing person ReID methods pay little attention to this resolution discrepancy problem.
One paradigm to deal with this problem is to use some complicated methods for mapping all images into an artificial image space, which however will disrupt the natural image distribution and requires heavy image preprocessing.
In this paper, we analyze the deficiencies of several widely-used objective functions handling image resolution discrepancies and propose a new framework called deep antithetical learning that directly learns from the natural image space rather than creating an arbitrary one.
We first quantify and categorize original training images according to their resolutions.
Then we create an antithetical training set and make sure that original training images have counterparts with antithetical resolutions in this new set.
At last, a novel Contrastive Center Loss(CCL) is proposed to learn from images with different resolutions without being interfered by their resolution discrepancies.
Extensive experimental analyses and evaluations indicate that the proposed framework, even using a vanilla deep ReID network, exhibits remarkable performance improvements.
Without bells and whistles, our approach outperforms previous state-of-the-art methods by a large margin.
\end{abstract}
%===========================================================

\section{Introduction}
Person re-identification (ReID) aims at identifying pedestrian identities across disjoint camera views.
It suffers from various difficulties such as large variations of pose, viewpoint, and illumination conditions.
Despite that person ReID tasks have been receiving increasing popularity, it remains a very challenging problem, especially in real-world application scenarios.

\begin{figure}[t]
  \begin{subfigure}[b]{0.44\textwidth}
    \includegraphics[width=\textwidth]{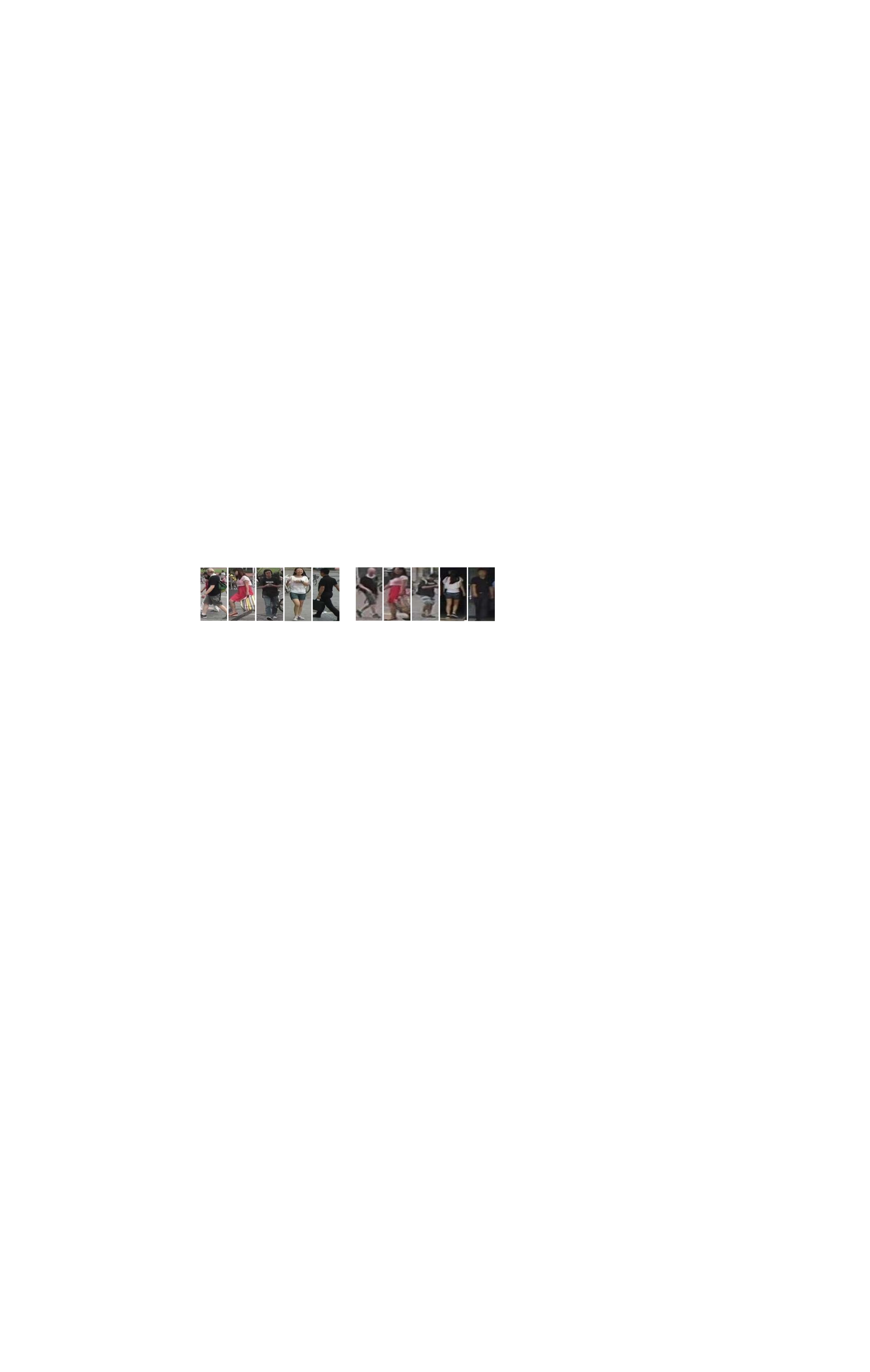}
    \caption{High resolution images}
  \end{subfigure}
  \hfill
  \begin{subfigure}[b]{0.44\textwidth}
    \includegraphics[width=\textwidth]{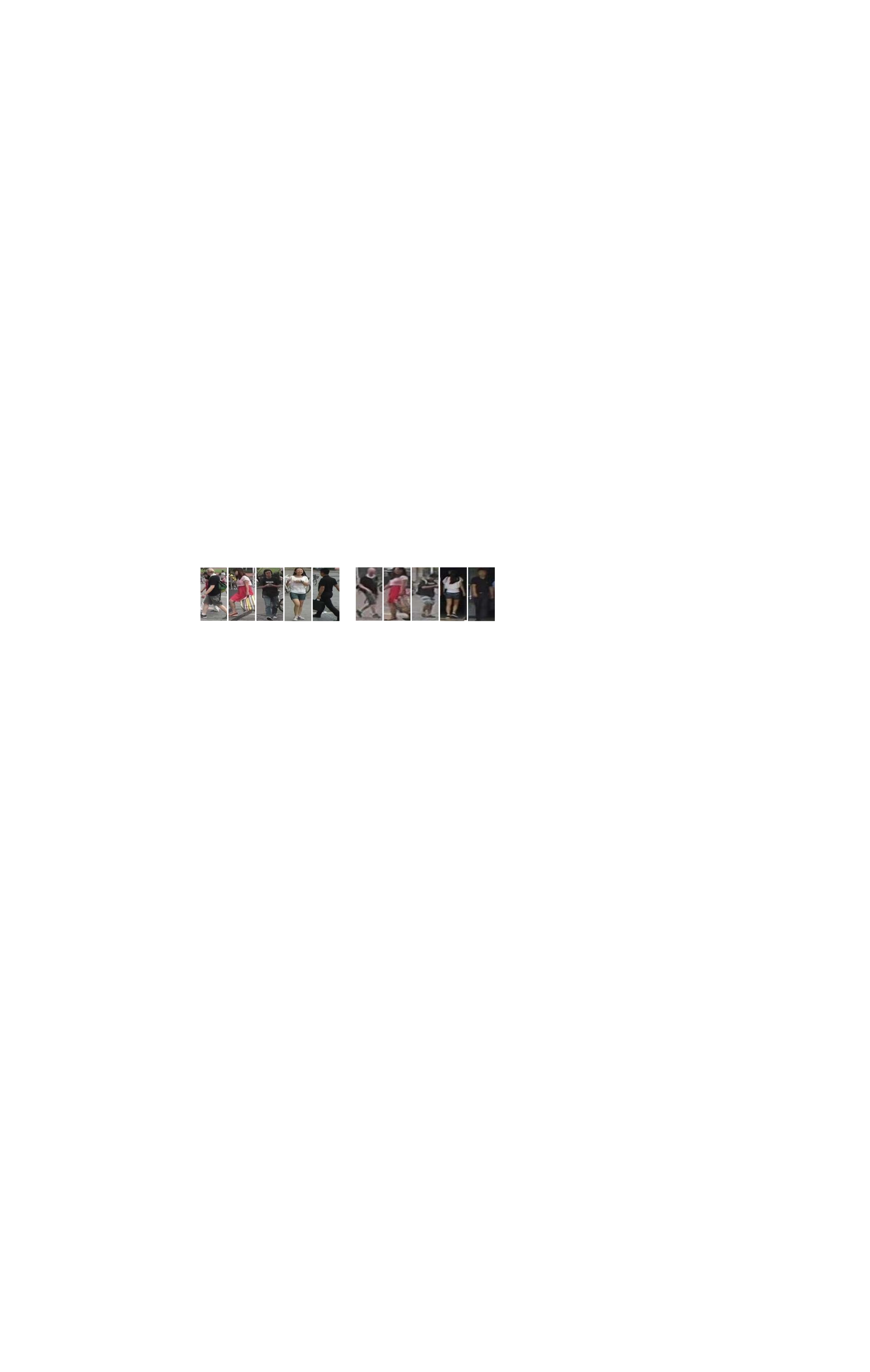}
    \caption{Low resolution images}
  \end{subfigure}
\caption{Examples of images with different resolution in public datasets.}
\label{fig:motivation}
\end{figure}

Recently, many inspiring works~\cite{FPNN,DeepAligned,BeyondPartModel,DPFL,NPSM} have been proposed to tackle issues such as part misalignment and viewpoint changes.
However, despite that these models have achieved remarkable performance on several person ReID benchmarks, two obvious, but as yet, unanswered questions are seldom valued by these approaches: 1) does the image resolution discrepancies in the training set affect the performance of person ReID? and 2) how to prevent a model from being prone to certain resolution combinations when the training data reflects the natural image distribution partially.
As shown in Fig.~\ref{fig:motivation}, the image resolution discrepancy problem is common in both public datasets and real-world applications.
We argue that these discrepancies are caused by arbitrarily rescaling training images with different resolutions to a uniform size.
The original resolutions of pedestrian image patches are diverse due to three reasons.
First, the graphical perspective leads to various sizes of pedestrians in images.
% For a stationary surveillance camera, a person who stands far away from the camera is naturally smaller than the one who is near to the camera, and image patches of smaller pedestrians have lower resolutions than that of larger pedestrians.
Second, configurations of surveillance cameras are different in both public datasets and real-world applications.
Some old surveillance cameras can only produce low-resolution images while other modern cameras generate high-resolution images.
% Therefore, when we crop pedestrian patches from the entire camera view, the resolutions of these image patches are diverse.
Third, to the best of our knowledge, almost all deeply-learned ReID models require rescaling image patches to a uniform size in both training and testing.
This procedure will inevitably lead to the image resolution discrepancy problem.

For a person ReID model, sufficient training data with different resolutions is vital for improving its generalization ability.
% Training a person ReID model with images with different resolution is vital for improving generalization ability.
For each image in the training set, if we get all its antithetical counterparts that have the same content but with different resolutions, it will help a ReID model to gain a better generalization ability.
However, there is almost no chance of finding a pair of images in which the image from the low-resolution camera has a higher image resolution than the one from the high-resolution camera.
It means that the resolution discrepancies in the actual training set are biased since certain resolution combinations are missing.
% Our experiments in section~\ref{COMPARE_LOSSES} demonstrate that these biased training samples affect the performance of the person ReID model significantly.
Previous methods cover up this problem with carefully designed training hyperparameters~\cite{MSML,InDefense,AlignedReID,BeyondPartModel,SVD} or sophisticated image pre-processing method~\cite{DLR}.
Unlike these methods, we propose a generic and straightforward framework called deep antithetical learning that directly tackles the resolution discrepancy problem.
The first step is the image quality assessment.
Since the resolution changes of training images are mostly caused by manually rescaling images into a uniform size, we adopt the No-reference Image Quality assessment (NR-IQA)~\cite{NR-IQA} and measure the image resolution in the frequency domain.
In the second step, we generate an antithetical training set in which the resolution of images is antithetical to their counterparts in the original training set.
Image counterparts of lower resolutions can be easily generated by randomly downsampling, while approaches for enhancing the image resolution are limited.
Generative adversarial networks (GANs) provide a practical approach for that purpose.
However, neither CycleGAN~\cite{CycleGAN} nor SRGAN~\cite{SRGAN} has the ability to enhance the image resolution to a specific level.
Despite that we can split the original training set into multiple subsets, we cannot guarantee that every image has counterparts in every subset.
Therefore, we roughly split the entire training set into two subsets: one with high-resolution (HR) images and another with low-resolution (LR) images.
We then generate an antithetical training set in which the resolution of images is antithetical to their counterparts in the original training set.
Specifically, for those HR images in the original set, we generate their LR counterparts by downsampling them randomly.
And for those LR images in the original set, a GAN-based model is utilized for recovering fine texture details from them.
These recovered images, along with the aforementioned manually blurred images, form the antithetical training set.

Apart from generating a new training set for better representing the natural image distribution, training the ReID model with proper objective functions is also crucial.
We analyze the widely-used identification+verification paradigm~\cite{IDENTITY+VERIFY} and find that the triplet loss with online hard negative mining (OHM) has a tendency to select training triplets of certain resolution combinations.
This selection bias makes the ReID model suffer from resolution discrepancies and severely damages the performance.
We address this problem by proposing a novel Contrastive Center Loss (CCL).
The intuition behind is that rather than designing a sophisticated strategy for handling resolution differences between positive image pairs and negative ones, it is much easier to consider positive samples and negative samples separately.
During the training procedure, the proposed CCL simultaneously clusters images of same identities and pushes the centers of different clusters away.
To summarize, our contribution is three-fold:
\begin{itemize}
\item We focus on the image resolution discrepancy problem, which is seldom valued by previous methods as far as we know. We propose a training framework that produces antithetical images from the original training set and utilizes these images to eliminate biased discrepancies during the training phase.
\item Unlike the previous super-resolution based ReID method~\cite{DLR}, the goal of the proposed framework is to accommodate actual images whose resolution is naturally various.
The proposed method does not require arbitrarily enhancing LR images during the test phase.
Therefore, it has a potential to serve as a practical method for boosting many existing ReID methods.
\item We go deep into the training procedure and investigate how the resolution discrepancies interfere with the triplet selection.
The proposed Constrastive Center Loss shows an ability to learn discriminative features from images regardless of their various resolutions.
\end{itemize}

% Based on the above contributions, 
In conclusion, we present a high-performance person ReID system.
Extensive experimental analyses and evaluations are conducted to demonstrate its effectiveness.
Without bells and whistles, the proposed approach outperforms previous state-of-the-art methods on three large benchmarks by a large margin.

\section{Related work}

% In this section, we introduce relevant works on the image quality assessment (IQA), generative adversarial networks (GAN), and person re-identification (ReID).

\textbf{Image Quality Assessment}.
Image quality assessment (IQA) is an important research area.
% It provides valuable information about visual degradation in an image and possible solutions for reconstruction.
It can be accomplished in three ways: full reference image quality assessment (FR-IQA), reduced reference image quality assessment (RR-IQA), and no reference image quality assessment (NR-IQA).
NR-IQA algorithms measure the quality of an image without the need for any reference image or its features.
Recently, various strategies have been proposed to measure image quality, including edge detection~\cite{EdgeDetection}, natural scene statistics~\cite{NSS}, wavelet decomposition~\cite{WAVELET1,WAVELET2}, and human visual system model~\cite{HVS}.
In this work, since the rescaling procedure is the major source of visual degradation, we evaluate the resolution of images with their sharpness.

\textbf{Generative Adversarial Network}.
Generative adversarial network (GAN) contains two sub-networks: a generator and a discriminator.
% The generator produces fake images while the discriminator attempts to distinguish them from real images.
The framework of GANs is first proposed by Goodfellow \etal~\cite{GAN}.
After that, many researchers focus on improving the stability and visual quality of GANs~\cite{DCGAN,CycleGAN,DeblurGAN}.
In the field of computer vision, GANs are widely used in applications ranging from motion deblurring (DeblurGAN)~\cite{DeblurGAN} to texture recovering (SRGAN)~\cite{SRGAN}.
To generate the antithetical training set, we adopt SRGAN~\cite{SRGAN} for recovering the fine texture details from low-quality images.

\textbf{Person Re-identification}.
Person re-identification (ReID) can be split into two subproblems: feature representations and distance metric learning.
Over the past decades, many studies focus on designing discriminative features~\cite{LBP,COLORNAMES1,COLORHIS1,COLORHIS2}, while others focus on constructing more robust metric learning algorithms~\cite{ITML,DML,KISSME}.
With the rise of deep learning, deeply-learned models have dominated person ReID tasks.
% FPNN~\cite{FPNN} is an early successful attempt to match the misaligned human parts with deep neural networks.
Several early works~\cite{FPNN,IDLA} take advantage of the two-stream siamese network and perform the pair-wise comparison in three steps: 1) extracting features from a given pair of images, 2) splitting feature cubes manually and comparing corresponding fractions across images, 3) determining whether these two images belong to the same identity.
Attention-based methods~\cite{NPSM,DeepAligned} provide a more adaptive way for locating different human parts.
% NPSM~\cite{NPSM} adopts the LSTM-based recurrent attention model to select and encodes salient areas iteratively.
% DeepAligned~\cite{DeepAligned} utilizes a simpler fully convolutional attention architecture and produces weighted masks directly from a certain convolutional layer. 
Unlike these methods which focus on handling the variations of human pose and viewpoint changes, the proposed method tackles another common but crucial problem: the biased image resolution discrepancies in the training data.

\section{Our approach}

\begin{figure}[!t]
\centering
\includegraphics[width=0.85\textwidth]{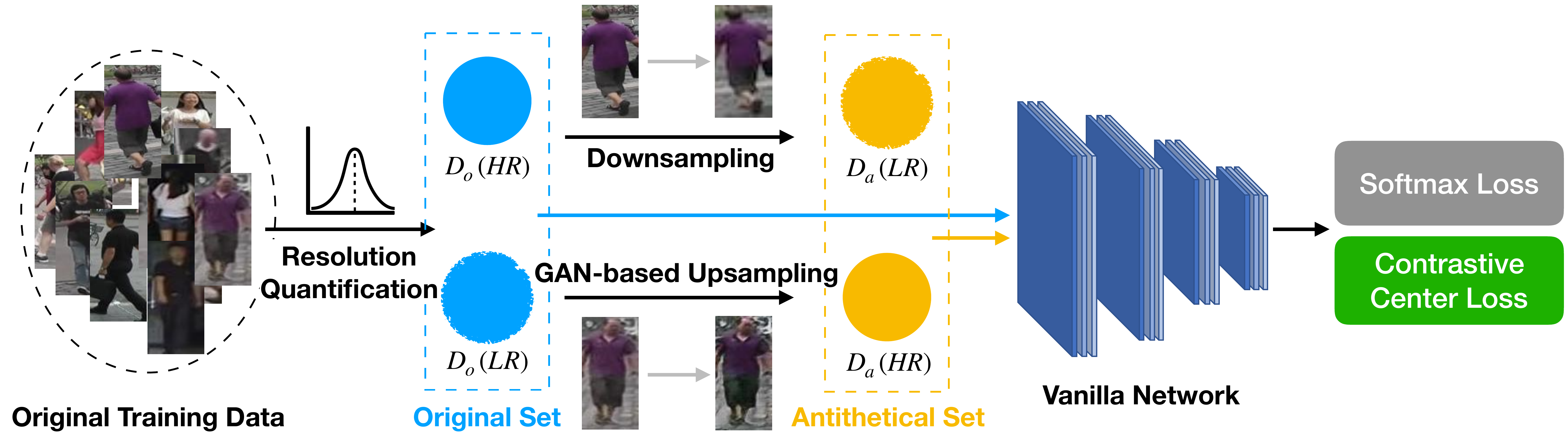}
\caption{
The proposed training pipeline.
}
\label{NETWORK}
\end{figure}

\subsection{Framework}

In person ReID tasks, the resolution of training images is naturally various.
However, previous methods seldom value these resolution discrepancies.
They probably learn biased mappings from these images.
% These networks fail to distinguish training images with different resolutions and learn biased mappings from these samples.
Besides, due to the fact that these discrepancies have significant impacts on distances between training images, some aggressive mining strategies such as online hard negative mining (OHM) will make the discrepancy problem even worse.
To deal with these issues, we propose an approach to train the person ReID model directly from these images by deep antithetical learning.
The motivations of the proposed deep antithetical learning are 1) producing antithetical training samples for balancing resolution discrepancies in the training set and 2) proposing a resolution-invariant objective function that produces better estimations of the image space.
As demonstrated in Fig.~\ref{NETWORK}, our approach mainly contains three steps.
First, we measure the resolution of each image in training set with the sharpness metric.
Second, we generate an antithetical training set by augmenting original low-resolution(LR) images with GANs and randomly downsampling original high-resolution(HR) images.
In this antithetical training set, the resolution of each image is antithetical to that of its counterpart in the original training set.
Third, after getting training samples from both the original training set and the antithetical set, we propose a novel Contrastive Center Loss (CCL) for 
learning relations between these images with various resolutions.

\subsection{Evaluation of Original Training Set}

The first step of generating the antithetical training set is to measure the resolution of images in the original training set.
Person ReID tasks have two significant characteristics.
% From the aspect of the data preprocessing procedure, 
1) The standard image preprocessing pipeline does not change the brightness or hue of images but only their resolutions.
% From the aspect of data, 
2) Images in ReID tasks are cropped image patches with tight bounding boxes, so the human body usually occupies a large portion of the entire image.
The abundant texture information from the identity appearance provides rich evidence for measuring the image blurriness.
We take advantage of the fact that sharper edges increase the high-frequency components and measure the resolution in the frequency domain.

We follow a simple sharpness metric proposed by Kanjar De \etal~\cite{MEASURESHARP}.
Given an image $I$ of size $h\times w$, we first compute its Fourier transform representation $F$.
Then we calculate the centered Fourier transform $F_{c}$ of image $I$ by shifting the origin of $F$ to center.
The threshold $\tau$ is defined as the maximum absolute value of $F_{c}$.
Now, we define the sharpness of an image $I$ as:

\begin{equation}
\vartheta(I)=\frac{1}{h \times w} \sum_{i=1}^{h}\sum_{j=1}^{w} \mathds{1}_{F_{i,j} \geq \left(\tau / 1000\right)} ,
\end{equation}
where $\mathds{1}_{\left(condition\right)}$ represents the indicator function.
After we obtain the sharpness of each image in the original training set $\it D_{o}$, we set up a threshold to split the entire set into two subsets $\it D_{o}(HR)$ and $\it D_{o}(LR)$.
The reason for only spliting the set $D_{o}$ into two subsets is that we lack the approach for tightly controlling the resolution of enhanced images.
Even if we split $D_{o}$ into multiple subsets, we cannot gurarantee that the resolution of enhanced LR images reaches a specific level.
We define this threshold as the mean sharpness of all images in the set $\it D_{o}$.
The subset $\it D_{o}(HR)$ contains images whose score is greater than this threshold, while images of inferior sharpness are collected into $\it D_{o}(LR)$.

\subsection{Antithetical Training Set}

As we mentioned above, images in the original training set $\it D_{o}$ are different not only in pose, viewpoint, illumination conditions but also in the image resolution.
Therefore, we propose to generate an antithetical training set $\it D_{a}$ for counteracting the biased resolution discrepancies.
In the previous section, we described how to quantify the image resolution and split the original training set into two subsets: $\it D_{o}(HR)$ and $\it D_{o}(LR)$.
Correspondingly, the antithetical training set $D_{a}$ also contains two subsets: $\it D_{a}(LR)$ and $\it D_{a}(HR)$.

For high-resolution images in the original subset $\it D_{o}(HR)$, the strategy for producing their antithetical low-resolution counterparts is straightforward.
For each image, we first downsample this image by a factor which is randomly chosen from a uniform distribution $\mathcal{U}(0.5,0.8)$, and then we rescale this image to its original size.
These manually blurred images are denoted as $\it D_{a}(LR)$.

For low-resolution images in the original training set $\it D_{o}(LR)$, we adopt SRGAN~\cite{SRGAN}, a GAN-based image super-resolution method, for recovering fine texture details from low-resolution images.
For each image of size $h\times w$ in $\it D_{o}(LR)$, SRGAN first upsamples it by a factor of $4$ and then rescales this image to its original size.
This rescaling procedure is necessary for eliminating random noises caused by SRGAN.
In this way, we obtain the antithetical high-resolution subset $\it D_{a}(HR)$.
For each low-resolution image in $\it D_{o}(LR)$, there is a corresponding high-resolution image in $\it D_{a}(HR)$.
We will give a detailed evaluation in Section.~\ref{QUANTIFY}.

\subsection{Contrastive Center Loss and Deep Antithetical Learning}

% In this section, we describe the proposed Contrastive Center Loss and the network of deep antithetical learning.

\begin{figure}[!htbp]
  \begin{subfigure}[b]{0.44\textwidth}
    \includegraphics[width=\textwidth]{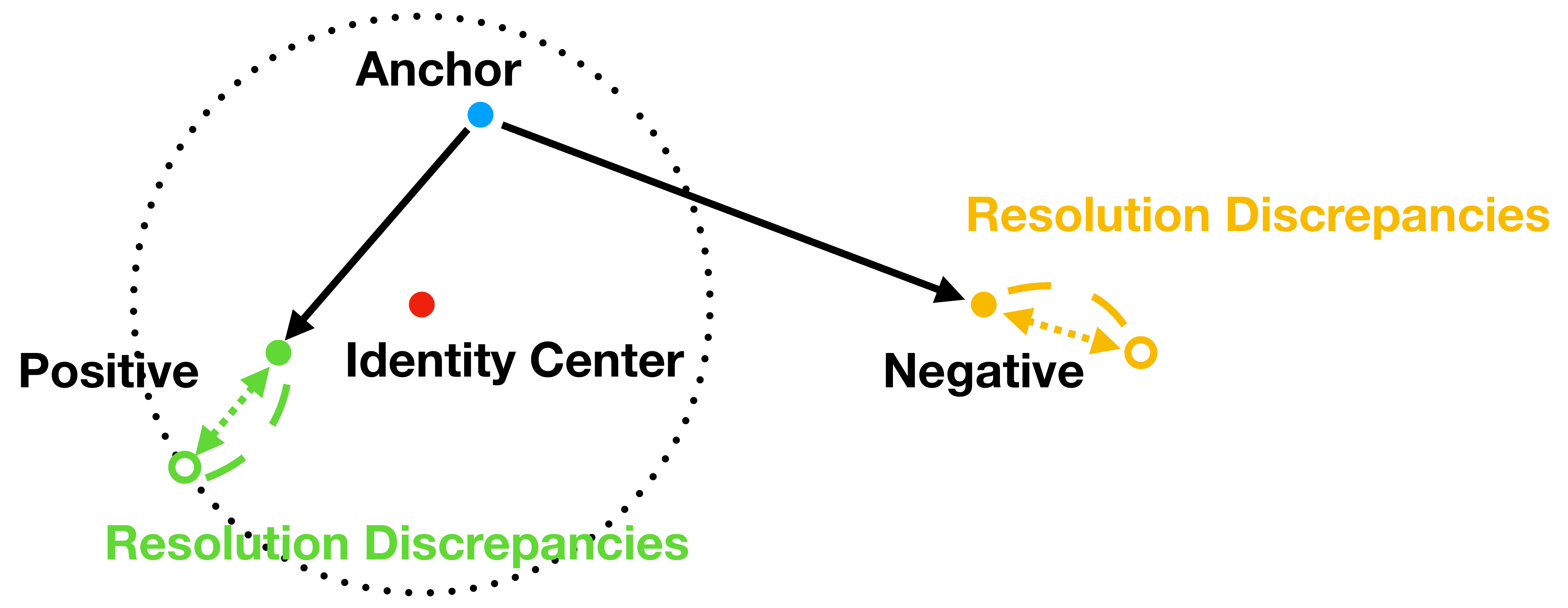}
    \caption{Visualization of Trihard}
    \label{fig:trihard_neglact}
  \end{subfigure}
  \hfill
  \begin{subfigure}[b]{0.54\textwidth}
    \includegraphics[width=\textwidth]{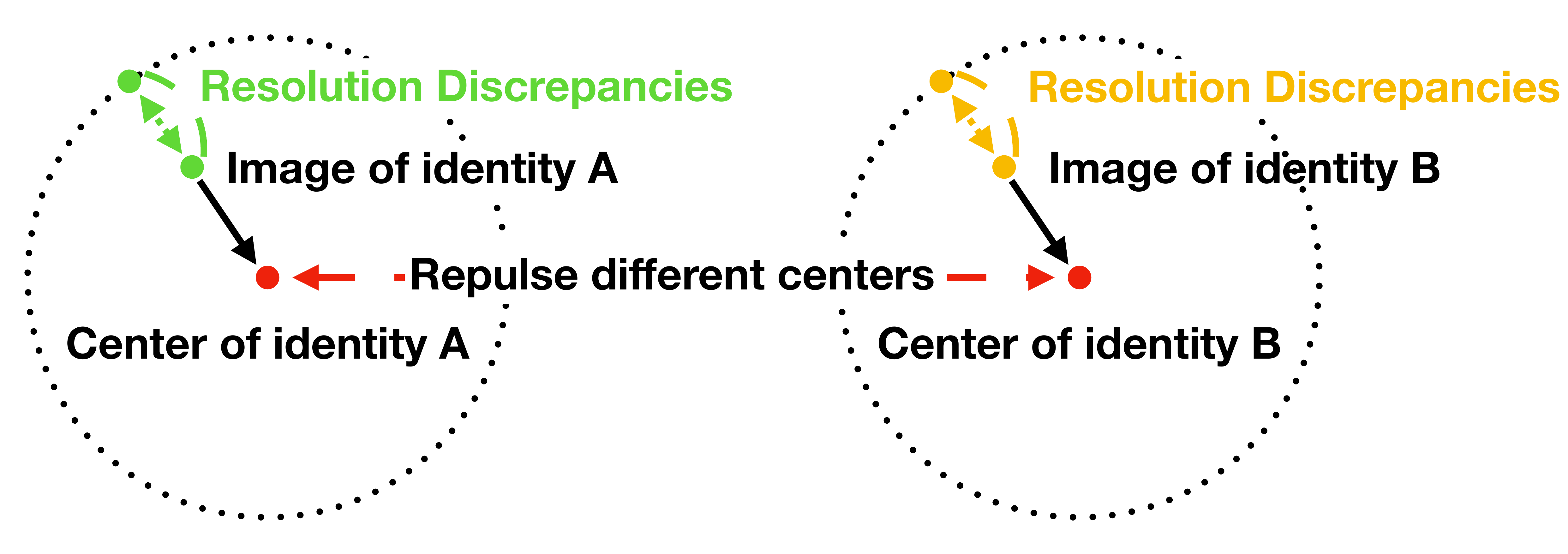}
    \caption{Visualization of CCL}
  \end{subfigure}
\caption{Differences between trihard and the proposed CCL.}
\label{fig:objective_func_difference}
\end{figure}

\textbf{Contrastive Center Loss}.
The proposed Contrastive Center Loss (CCL) aims at estimating the distance between different images without being interfered by their resolution discrepancies.
The softmax loss + triplet loss with online hard negative mining (trihard) approach is widely used in recent works.
This paradigm prefers positive images with the maximum distance to the anchor and negative images with the minimum distance to the anchor.
However, this paradigm neglects the fact that the resolution discrepancies have a salient influence on these distances(Fig.~\ref{fig:trihard_neglact}).
% As shown in Fig.~\ref{fig:pick_histogram}, 
We find that in the actual training procedure, trihard tends to select positive image pairs of which the resolution is most different, and negative image pairs of which the resolution is most similar.
This biased tendency keeps a ReID model trapped into the local optima and damages its generalization ability.
We will give a more detailed analysis in Section.~\ref{COMPARE_LOSSES}.

Fig.~\ref{fig:objective_func_difference} shows the difference between the proposed CCL and the trihard loss.
The trihard loss and the proposed CCL are both based on measuring distances between training samples.
As we mentioned before, resolution discrepancies have a significant impact on these distances.
For the trihard loss, resolution discrepancies in both positive samples and negative samples will affect the results of the loss function.
To reduce the negative influence of resolution discrepancies, the proposed CCL measures intra-identity distances and inter-identity distances separately.
For images of the same identity, we first estimate the center of each identity iteratively and minimize the distances between its center and corresponding image features: 

\begin{equation}
\mathcal{L}_{intra}= \frac{1}{N}\sum_{i=1}^{N}\left(1 - \cos \left(f_{i}, C_{y_{i}}\right)\right),
\end{equation}
where $f_{i}$ denotes the feature extracted from the $i$th image and $C_{y_{i}}$ stands for the corresponding center.
In this way, all features only connect to their corresponding identity centers, so that resolution discrepancies will not spread across different identities.
For inter-identity distances, to make the most of negative images and avoid perturbations caused by resolution variations, we use negative samples indirectly.
The relations of images of different identities are measured by the cosine distance of their corresponding centers.
Since maximizing cosine distances is equivalent to minimizing their cosine similarities, the loss for repulsing different centers is defined as:

\begin{equation}
\mathcal{L}_{inter}= \frac{1}{N}\sum_{i=1}^{N}\sum_{j=1}^{N}\abs{\cos \left(C_{y_{i}}, C_{y_{j}}\right)},
\end{equation}
where $\abs{\cdot}$ stands for the absolute value symbol.
The reason for using the absolute value is that the orthogonality relation between identity centers is more discriminative than the positive/negative correlation.
Note that both the intra-identity losses and the inter-identity losses are measured with cosine distances.
An advantage of the cosine metric is that its range is certain.
As shown in Table~\ref{distance_compare}, inter-identity Euclidean distances are much greater than intra-identity distances.
And during the training procedure, these two kinds of distances change at different speeds, and their corresponding losses change as well.
Since these two losses both rely on the trainable identity centers, it is important to keep them in a certain range. 
Finally, the Contrastive Center Loss (CCL) is formulated as:
\begin{equation}
\mathcal{L}_{LCC}= \alpha\mathcal{L}_{intra} + \beta\mathcal{L}_{inter}.
\end{equation}
The weight $\alpha$ and $\beta$ for balancing losses will be discussed in Section.~\ref{COMPARE_LOSSES}.

\textbf{Network Architecture}.
We now describe the network for deep antithetical learning.
With the help of the antithetical training set and the Contrastive Center Loss, even a vanilla deep network can achieve remarkable performance.
% We test the proposed framework with a vanilla deep network architecture.
% The network architecture is demonstrated in Fig.~\ref{NETWORK}.
We denote this deep network as ``VanillaNet'' in following sections.
VanillaNet contains two basic components: 1) a convolutional network backbone with a global average pooling layer for extracting features, 2) two successive fully-connected layers denoted as $FC_{0}$ and $FC_{1}$, where $FC_{1}$ is used for ID classification.
We use the standard Identification+Verification framework for training the VanillaNet.
The cross entropy loss of ID classification can be formulated as follow:

\begin{equation}
\mathcal{L}_{s}\left(p, g \right) = -\sum_{k=1}^{K}\operatorname{log}(p_{k})\mathds{1}_{k = g},
\end{equation}
where $\mathds{1}_{\left(condition\right)}$ is the indicator function.
$p$ represents the prediction and $g$ stands for the ground truth ID.
The proposed CCL is connected to the last ReLU layer of the CNN backbone and the overall objective function for the proposed framework is formulated as:
\begin{equation}
\mathcal{L} =  \mathcal{L}_{s} + \alpha\mathcal{L}_{intra} + \beta\mathcal{L}_{inter}.
\end{equation}
During the training phase, we simply feed images from both original and antithetical training set into the network.
In the testing phase, features from the last ReLU layer are used for ranking.

\section{Experiments}

\subsection{Datasets}

To evaluate the proposed framework, we select three public datasets: Market-1501~\cite{MARKET}, Duke-MTMC-reID~\cite{DUKEMTMCREID}, and CUHK03~\cite{CUHK03}.
% with both ``labeled'' and ``detected'' subsets.

\textbf{CUHK03.}
The CUHK03 dataset contains $14096$ images of $1467$ identities.
There are at most ten images for each identity shot by two disjoint cameras.
Unlike previous testing protocol which only adopts $100$ identities for testing, we follow the new protocol presented by Zhong \etal~\cite{RERANK}.
This new protocol adopts $767$ identities for training and the rest $700$ identities for testing.
Under this protocol, each identity has more than one ground truth image in the gallery, which is more consistent with the real-world applications.

\textbf{Market-1501.}
The Market-1501 dataset is a large ReID dataset which contains $32643$ annotated boxes of $1501$ different identities.
We divide this dataset into a training set of 750 identities and a testing set of 751.
Since images in this dataset are collected by the pedestrian detector, it involves several detector failures.
Besides, the quality of images shot by one particular camera is significantly lower than the quality of other images.
These two characteristics make this dataset suitable for quantifying the effectiveness of the proposed method.

\textbf{Duke-MTMC-reID.}
Duke-MTMC-reID is a newly published dataset.
It contains 36411 bounding boxes shot by 8 different cameras.
We use 16522 training images of 702 identities, leaving 2228 query images of the other 702 identities and 17661 gallery images for the testing procedure.
Unlike Market-1501, the quality of images in this dataset is much higher and more consistent.

\subsection{Implementation Details}

For the SRGAN, we use the same training parameters provided in its original paper~\cite{SRGAN}.
We first train the model on the DIV2K dataset and then fine-tune it on HR images from the ReID training set.
In the training phase, rather than cropping training images randomly, we directly pad HR training images with zeros and resize them to the target scale.

For the VanillaNet, we adopt the ResNet-50~\cite{ResNet} backbone in all experiments.
% For the VanillaNet, we evaluate three convolutional backbones: ResNet-50~\cite{ResNet}, ResNet-101, and ResNeXt-50~\cite{ResNeXt}.
% We compare the performance of VanillaNet with and without the proposed LSN under the same configuration.
The batch size is set to $60$.
% Note that when training VanillaNet with LSN, there are only images of $30$ identities since the antithetical set provides images of the same identities as the original training set.
% The weight $\alpha$ for balancing two different objective functions is set to $1.0$ initially and decays by a factor of $0.1$ whenever the classification loss is less than the similarity loss.
Both the weight $\alpha$ and $\beta$ are set to $0.1$ and the output dimension of the CNN backbone is $2048$.
In both training and testing phase, all images are resized to the size of $256\times 128$.
The data augmentation includes RandomErasing~\cite{RANDOMERASE} and random horizontal flipping.
To train the model, we adopt stochastic gradient descent (SGD)~\cite{SGD} optimizer with an initial learning rate of $0.01$ and weight decay of $5\times 10^{-4}$.
In all experiments, the training phase lasts for 60 epochs.
And the learning rate starts to decay exponentially at 20th epoch with the base of $0.1$.
The overall time cost of training the proposed model is minor.
For the Market-1501, it takes about 130 minutes for the training procedure on a single GTX-Titan-Xp GPU.

\subsection{Quantifying Image Resolution}
\label{QUANTIFY}

\begin{figure}[h]
\centering
\includegraphics[width=0.9\textwidth]{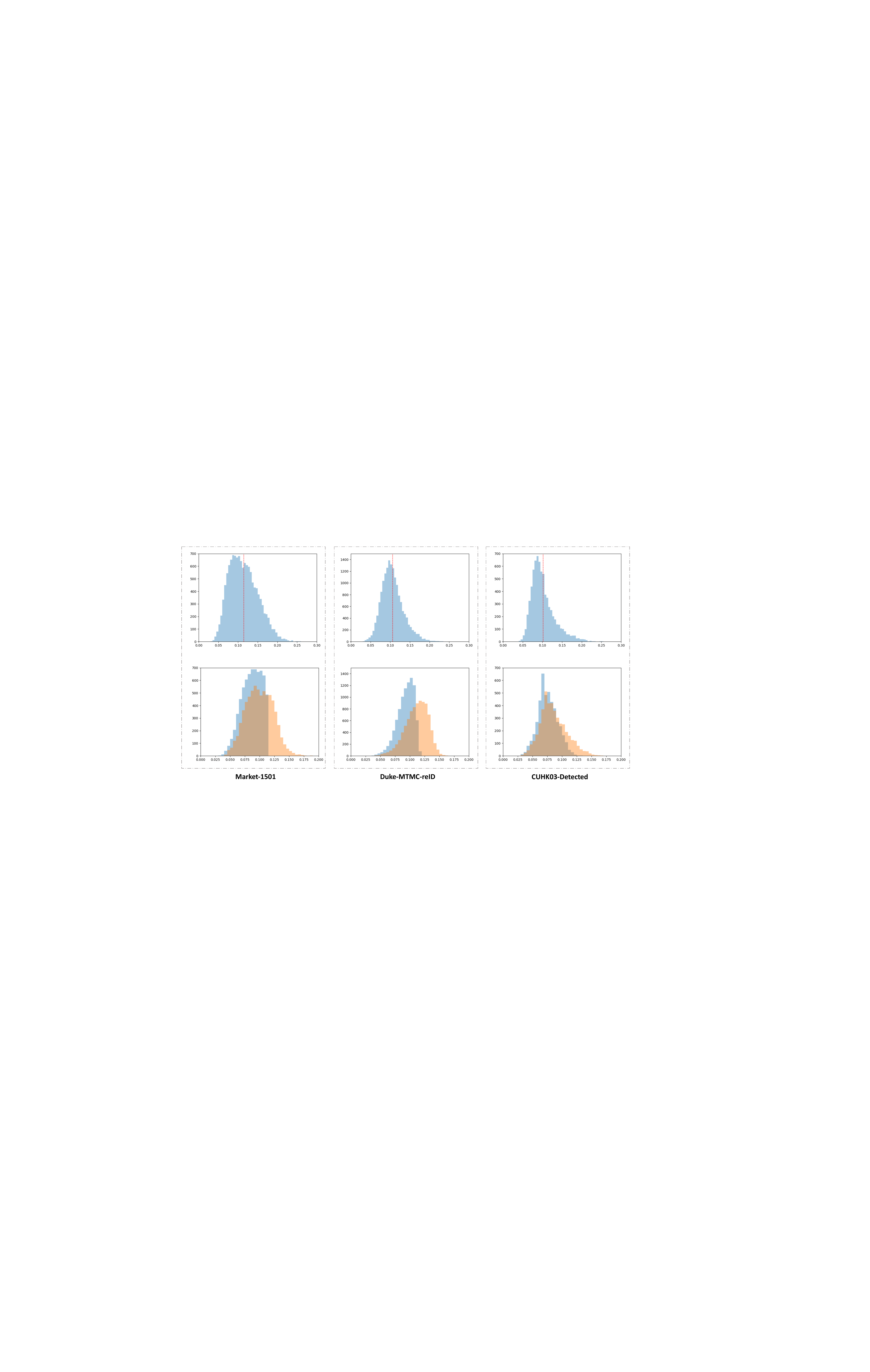}
\caption{
Visualizations of the image resolution in different datasets.
% Each diagram in the first row is the sharpness histogram of the original training set $\it D_{o}$.
% And each diagram in the second row compares the histogram of low-resolution subset $\it D_{o}(LR)$ (marked with blue) and its antithetical counterpart $\it D_{a}(HR)$ (marked with orange).
}
\label{fig:resolution_hist}
\end{figure}

\begin{figure}[h]
  \begin{subfigure}[b]{0.45\textwidth}
    \includegraphics[width=\textwidth]{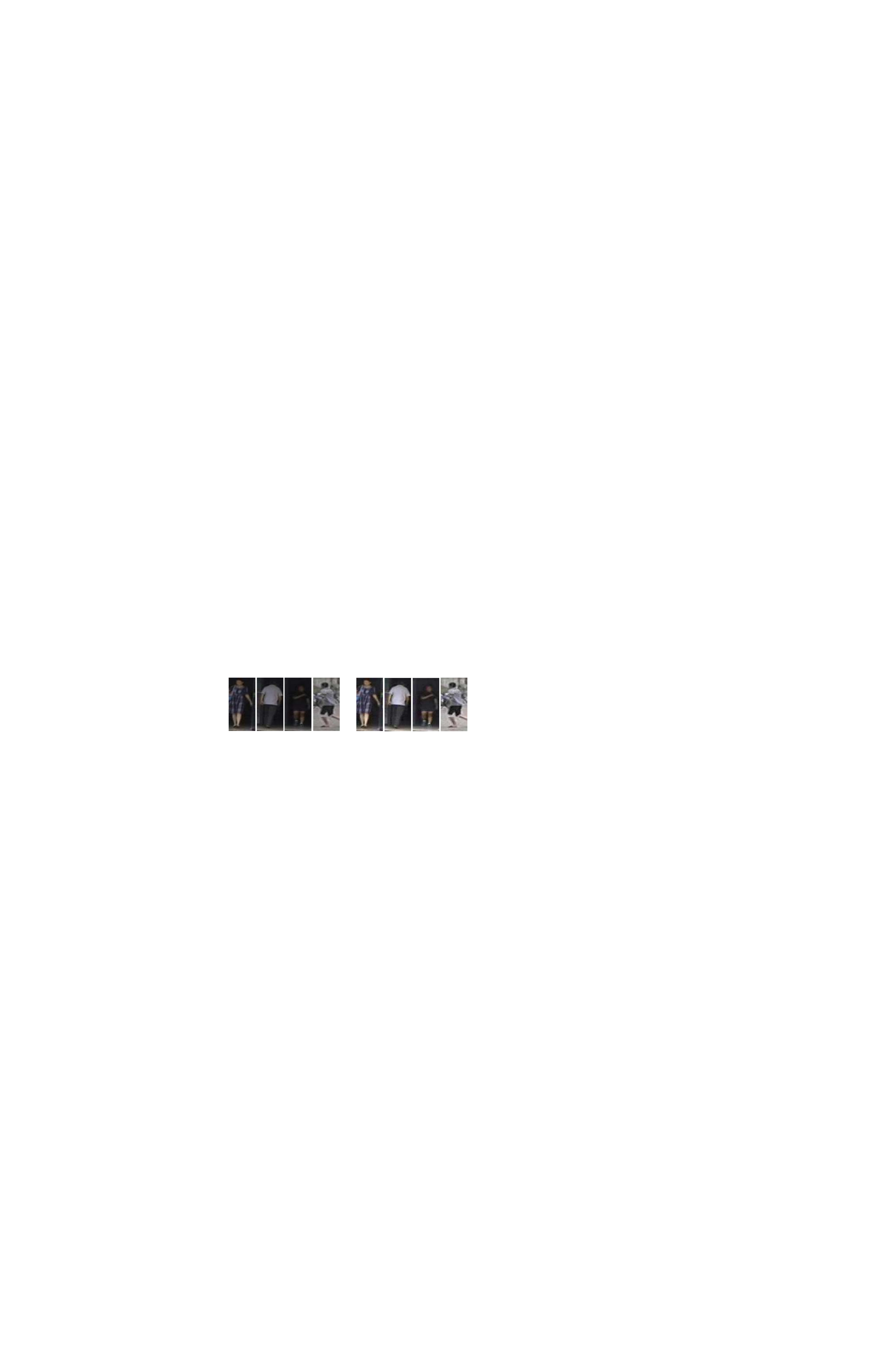}
    \caption{Original LR images}
  \end{subfigure}
  \hfill
  \begin{subfigure}[b]{0.45\textwidth}
    \includegraphics[width=\textwidth]{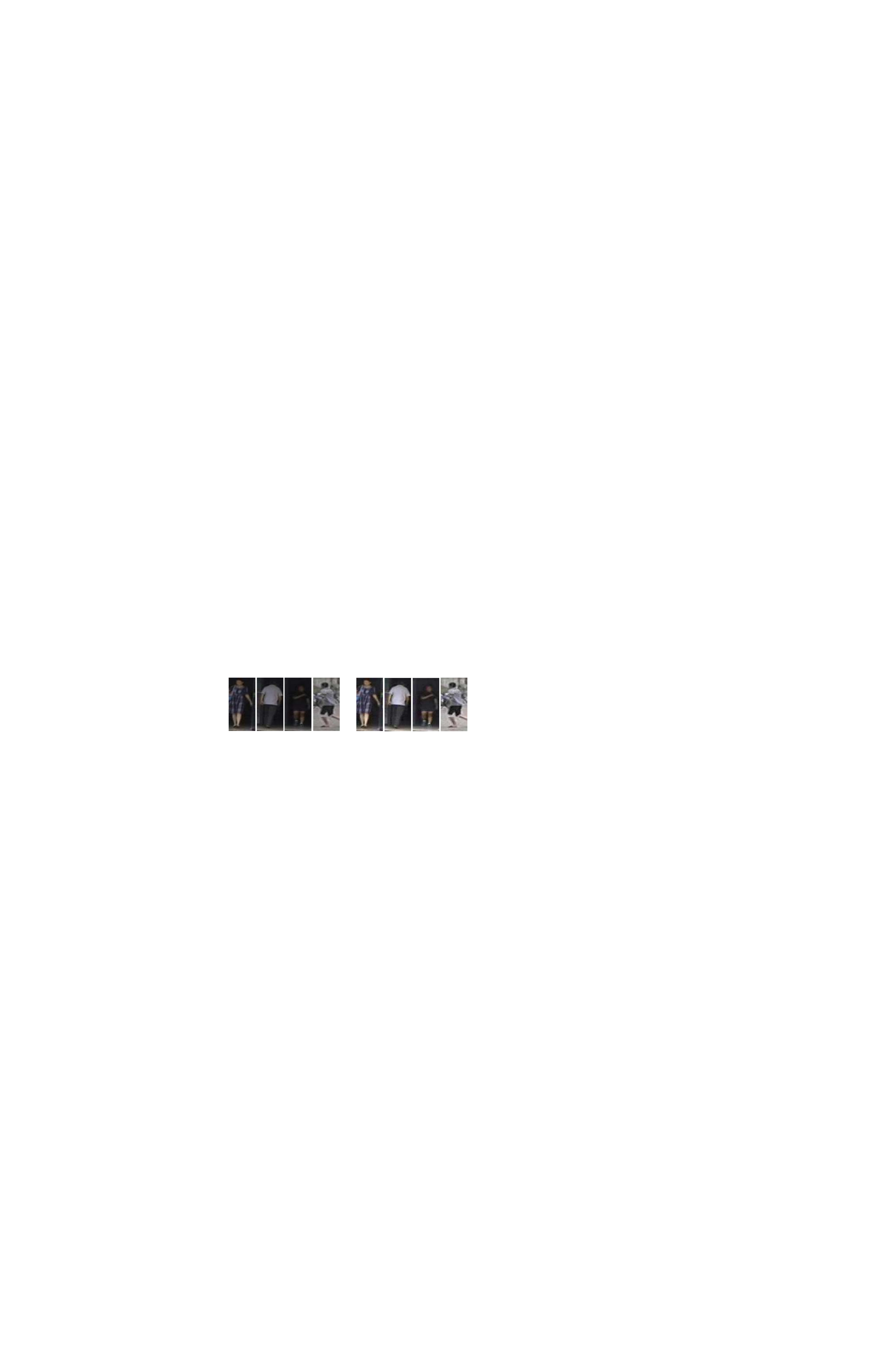}
    \caption{Antithetical HR images}
  \end{subfigure}
\caption{
Examples of low-resolution images in the original training set and their counterparts in the antithetical set.
% Compared to original images, improvements in texture details of the clothing, boundaries between foreground and background, and the brightness of the human body under low light conditions are notable.
}
\label{fig:augment}
\end{figure}

In this section, we analyze image resolution distributations of all datasets.
Since two subsets of CUHK03 are similar in human pose, viewpoint, and illumination conditions, we only present histograms of the ``detected'' subset.
As shown in Fig.\ref{fig:resolution_hist}, diagrams in each column correspond to Market-1501, Duke-MTMC-reID, and CUHK03 (detected), respectively.
Histograms in the first row represent the resolution distributation of the original training set $\it D_{o}$.
And the red dashed line in each of them is the threshold for splitting the original set.
In the second row, we compare the resolution of images in the original LR set $\it D_{o}(LR)$ and its corresponding antithetical set $\it D_{a}(HR)$.
The blue histogram in each diagram corresponds to $\it D_{o}(LR)$, and the orange one corresponds to $\it D_{a}(HR)$.
We also present the statistical analyses in Table~\ref{QUANTIFY_DATA} and some examples in Fig.~\ref{fig:augment}.

In summary, for Market-1501, Duke-MTMC-reID, and CUHK03 (labeled), SRGAN can significantly augment low-resolution images in the original training set, especially under low light conditions. 
% We also observe that, compared to these three set, the improvements on CUHK03 (detected) is not conspicuous.

% \usepackage{multirow}
\begin{table}[h]
\centering
\begin{tabular}{c|C{1.3cm}C{1.3cm}C{1.3cm}|C{1.3cm}C{1.3cm}C{1.3cm}}
\hline
\multirow{2}{*}{\textbf{Datasets}} & \multicolumn{3}{c|}{\textbf{original set $\it D_{o}(LR)$}}  & \multicolumn{3}{c}{\textbf{antithetical set $\it D_{a}(HR)$}}                                 \\ \cline{2-7} 
                          & \textbf{num}      & \textbf{mean}  & \textbf{median} & \textbf{num}     & \textbf{mean}     & \textbf{median} \\ \hline \hline
Market-1501               & \multicolumn{1}{c|}{6906} & \multicolumn{1}{c|}{0.087} & 0.091  & \multicolumn{1}{c|}{6906} & \multicolumn{1}{c|}{0.100} & 0.102  \\ \hline
Duke-MTMC-reID            & \multicolumn{1}{c|}{9121} & \multicolumn{1}{c|}{0.093} & 0.075  & \multicolumn{1}{c|}{9121} & \multicolumn{1}{c|}{0.111} & 0.089  \\ \hline
CUHK03 (labeled)           & \multicolumn{1}{c|}{4661} & \multicolumn{1}{c|}{0.075} & 0.088  & \multicolumn{1}{c|}{4661} & \multicolumn{1}{c|}{0.085} & 0.101  \\ \hline
CUHK03 (detected)          & \multicolumn{1}{c|}{4444} & \multicolumn{1}{c|}{0.076} & 0.066  & \multicolumn{1}{c|}{4444} & \multicolumn{1}{c|}{0.086} & 0.070  \\ \hline
\end{tabular}
\caption{
Quantifying the mean and median of image sharpness scores before and after enhancements.
Both the mean and median sharpness scores of low-resolution images are improved.
% The mean and median of sharpness scores before and after enhancements are presented.
}  % of $\it D_{o}(LR)$ and $\it D_{a}(HR)$}
\label{QUANTIFY_DATA}
\end{table}

\subsection{Analyzing Different Data Fusion Strategies}
\label{ANALZE_DATA_MIX}

The antithetical training set $D_{a}$ is produced by two different approaches: enhancing LR images with SRGAN and downscaling HR images randomly.
This specific strategy seems unsymmetrical.
In this section, we demonstrate that both strategies are crucial for better estimating the real-world distribution.
All following experiments are conducted on the test set of Market-1501.
We first evaluate whether the antithetical training set improves the ReID performance.
Only softmax identification loss is applied to VanillaNet in these experiments.
As shown in Table~\ref{training_data}, both the enhanced set $D_{a}(HR)$ and the decayed set $D_{a}(LR)$ are beneficial to the ReID performance.
When combining $D_{a}(HR)$, $D_{a}(LR)$, and $D_{o}$ together, VanillaNet reaches the highest performance.

\begin{table}[h]
\RawFloats
\centering
\makebox[0pt][c]{\parbox{1\textwidth}{
\begin{minipage}[b]{0.5\hsize}\centering
   \begin{tabular}{C{2.0cm}|C{1.2cm}|C{1.2cm}}
      \hline
      \textbf{Datasets}   & \textbf{rank-1} & \textbf{mAP}   \\ \hline \hline
      $D_{o}$             & 88.63  & 72.47 \\ \hline
      $D_{o}$+$D_{a}(LR)$ & 89.16  & 73.98 \\ \hline
      $D_{o}$+$D_{a}(HR)$ & 89.84  & 73.75 \\ \hline
      $D_{o}$+$D_{a}$     & \textbf{90.11}  & \textbf{74.33} \\ \hline
   \end{tabular}
   \caption{Comparing training data.}
   \label{training_data}
\end{minipage}
\hfil
\begin{minipage}[b]{0.5\hsize}\centering
   \begin{tabular}{C{1.0cm}|C{1.1cm}|C{1.1cm}|C{1.1cm}|C{1.1cm}}
      \hline
      \multirow{2}{*}{\textbf{Probe}} & \multicolumn{2}{c|}{$\boldsymbol{D_{o}}$} & \multicolumn{2}{c}{$\boldsymbol{D_{o}}$+$\boldsymbol{D_{a}}$} \\ \cline{2-5} 
                             & \textbf{rank-1}       & \textbf{mAP}      & \textbf{rank-1}        & \textbf{mAP}       \\ \hline\hline
      LR                             & 85.48         & 68.95        & 87.25             & 70.86            \\ \hline
      HR                             & 92.26         & 76.54        & 93.41             & 78.33            \\ \hline
      ALL                            & 88.63         & 72.47        & 90.11             & 74.33    \\ \hline
   \end{tabular}
   \caption{Comparing query probe.}
   \label{probe_result}
\end{minipage}
\hfil
}}
\end{table}

Following the same criterion for splitting the training set, we further divide all query images into high-resolution probes and low-resolution probes.
As shown in Table~\ref{probe_result}, the performance of querying with LR probes is much lower than that of querying with HR probes.
Furthermore, we notice significant improvements in both LR queries and HR queries when adopting the antithetical training set.
These results indicate that the ReID model benefits from not only the SRGAN but also the random downsampling procedure.
To further prove this conclusion, we compare the performance of our data augmentation approach with the other two approaches: 1) enhancing all images in $D_{o}$ and 2) downsampling all these images.
Table~\ref{fusion_strategy} demonstrates the performance of these approaches and internal differences on image distances.
Note that these experiments are conducted with both the softmax loss and the proposed CCL for clustering images and tracking the identity centers.
$D_{intra}$ and $D_{inter}$ stand for the average distance between images of the same identity, images of different identities on the test set.
$D_{centers}$ represents the average distance between all identity centers on the training set.
We adopt $D_{centers}$ for measuring the separation of different identity clusters.

Compared to other two fusion strategies, VanillaNet with the proposed CCL and antithetical set obtains the smallest intra-identity distances and the largest center distances.
These results indicate that VanillaNet gains a better generalization ability on the test set with the proposed CCL and antithetical images.

\begin{table}[h]
\centering
\begin{tabular}{C{4.5cm}|C{1.3cm}|C{1.3cm}||C{1.3cm}|C{1.3cm}||C{1.3cm}}
\hhline{-|-|-||-|-||-}
\textbf{Data Fusion Strategy}    & \textbf{rank-1} & \textbf{mAP}   & $\boldsymbol{D_{intra}}$     & $\boldsymbol{D_{inter}}$     & $\boldsymbol{D_{centers}}$    \\ \hhline{=|=|=||=|=||=}
Original+All SRGAN     & 89.85  & 74.66 & 0.4559                             & 0.8075             & 0.6100           \\ \hhline{-|-|-||-|-||-}
Original+All Downscale  & 87.14  & 72.47 & 0.4590                             & \textbf{0.8257}    & 0.6085          \\ \hhline{-|-|-||-|-||-}
Original+Our Approach  & \textbf{90.83}  & \textbf{76.63} & \textbf{0.4548}  & 0.8127             & \textbf{0.6122}  \\ \hhline{-|-|-||-|-||-}
\end{tabular}
\caption{Comparing fusion strategies ($\alpha=0.1$, $\beta=0.1$).}
\label{fusion_strategy}
\end{table}

\subsection{Comparing CCL with Other Objective Functions}
\label{COMPARE_LOSSES}

In this section, we will discuss the differences between triplet loss with OHM (trihard), Center Loss, and the proposed Contrast Center Loss(CCL).
% With proper warmming-up strategies, Online Hard Negative Mining(OHM) boosts the effeciency and performance of naive triplet loss by constantly selecting the hardset training samples.

\begin{figure}[!htbp]
  \begin{subfigure}[b]{0.45\textwidth}
    \includegraphics[width=\textwidth]{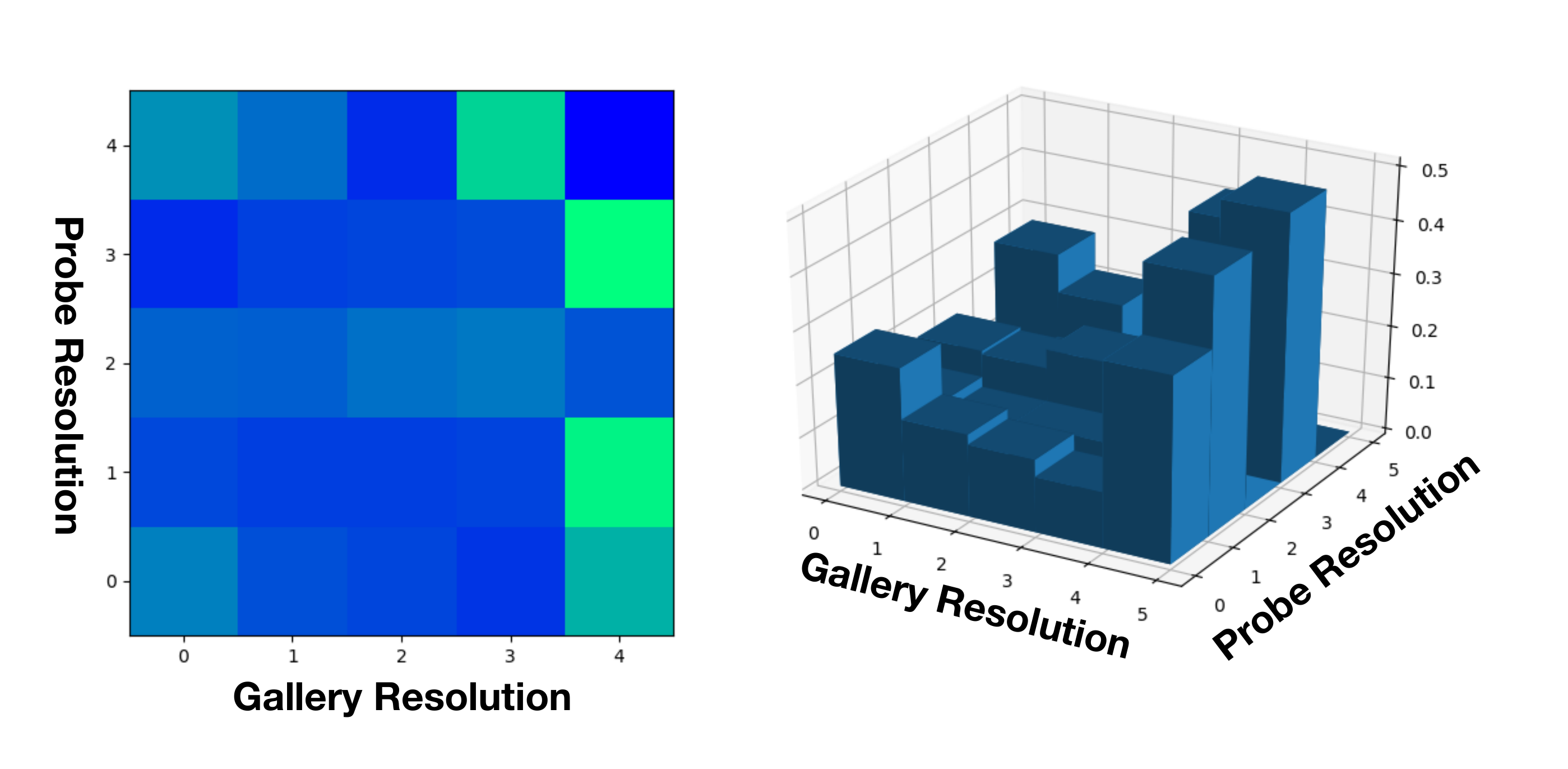}
    \caption{Positive pairs}
  \end{subfigure}
  \hfill
  \begin{subfigure}[b]{0.45\textwidth}
    \includegraphics[width=\textwidth]{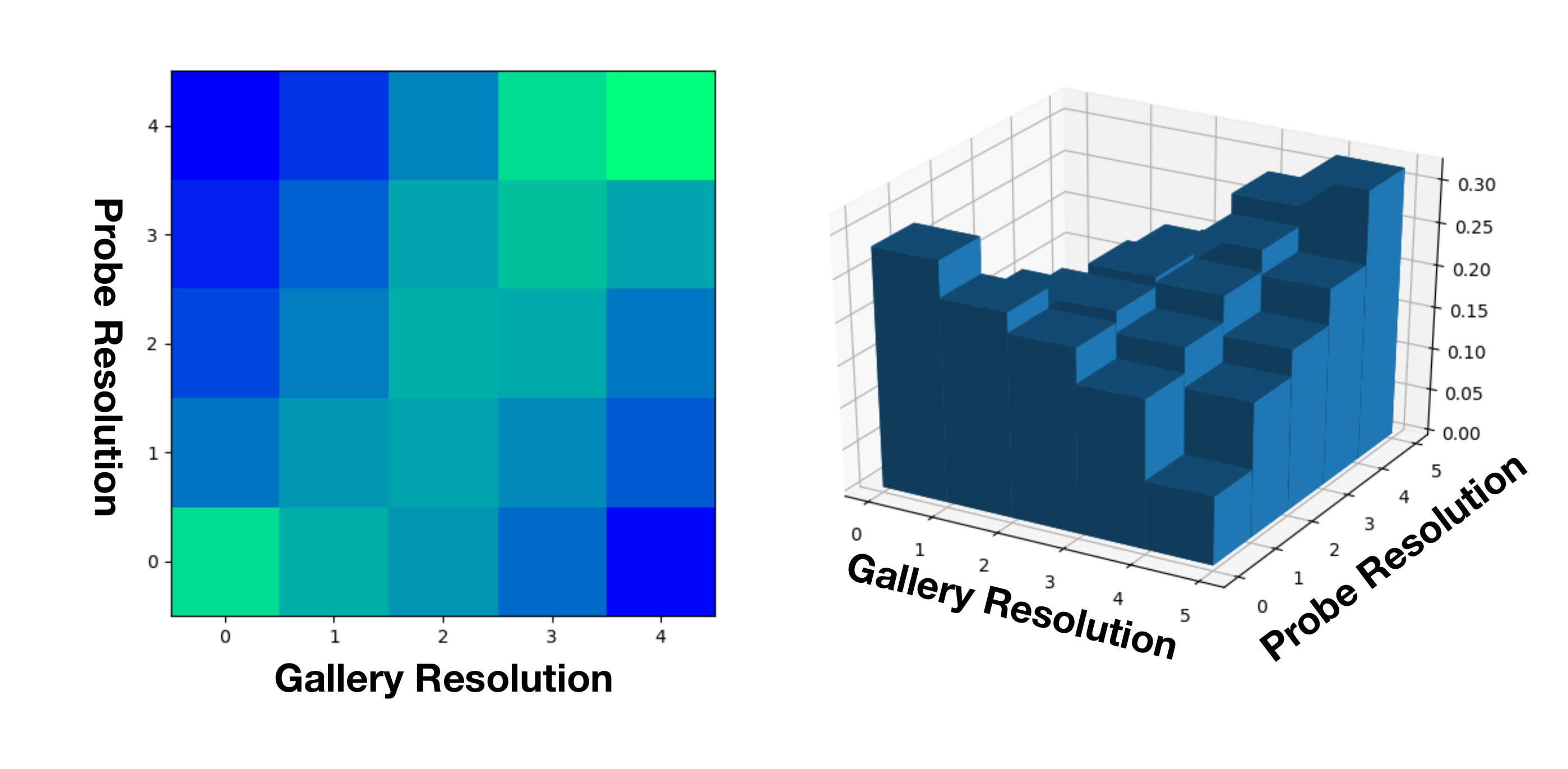}
    \caption{Negative pairs}
  \end{subfigure}
\caption{
The selection tendency of TriHard.
Picked positive image pairs usually have the biggest difference in resolution, while the negative pairs have the most similar resolution.
These histograms are normalized along the gallery axis.
}
\label{fig:pick_histogram}
\end{figure}

For trihard, our experiments indicate that the resolution of training images has biased influences on the triplet-picking procedure.
Given a probe image, trihard expects the farthest image of the same identity and the nearest image of a different identity.
As shown in Table~\ref{distance_compare}, when the probe is with low-resolution, it is more likely for trihard to pick a positive HR image and a negative LR image at the same time.
% In a word, it is harder for trihard to learn from image triplets of similar resolutions.
We also track all selected triplets during the training phase.
Histograms in Fig.~\ref{fig:pick_histogram} show the possibility of picking image pairs with certain resolution combinations.
% all picked probe-positive and probe-negative image pairs.
The selected positive images tend to have a most different resolution than that of probe images, while the resolution of picked negative images tends to be the same as that of probe images.
In a word, trihard suffers from resolution discrepancies and fails to learn all possible image combinations.

\begin{table}[h]
\centering
\begin{tabular}{c|c|c}
\hline
\textbf{Distance between selected image pairs}    & \textbf{intra-identity} & \textbf{inter-identity} \\ \hline\hline
(sharp, sharp)                                 & 0.2580                  & 0.7306         \\ \hline
(blur, blur)                                      & 0.2475                  & 0.7182         \\ \hline
(sharp, blur) and (blur, sharp)    & 0.2754                  & 0.7255         \\ \hline
\end{tabular}
\caption{Distances between images with different resolutions (Training Set).}
\label{distance_compare}
\end{table}

For the Center Loss, the resolution discrepancy problem is much less severe.
Images are only used for estimating their corresponding identity centers, so the discrepancies will not spread across different identities.
However, it is at the cost of ignoring all negative images.
% The proposed Contrastive Center Loss (CCL) overcomes this flaw by proposing 
Unlike Center Loss, the proposed CCL manages to learn from negative samples indirectly.
When updating identity centers, the proposed CCL not only reduces the distance between image features and their corresponding center but also pushes different centers away.
In this way, images are connected to their relevant centers directly and irrelevant centers indirectly.
As shown in Table~\ref{compare_cl_ccl}, the proposed CCL significantly increases the distances between different centers and distances between images of different identity.
At the same time, the average intra-identity distance is slightly larger.

\begin{table}[!h]
\centering
\begin{tabular}{C{5.0cm}|C{1.3cm}|C{1.3cm}||C{1.3cm}|C{1.3cm}||C{1.3cm}}
\hhline{-|-|-||-|-||-}
\textbf{Data Fusion Strategy} & \textbf{rank-1} & \textbf{mAP}   & $\boldsymbol{D_{intra}}$ & $\boldsymbol{D_{inter}}$ & $\boldsymbol{D_{centers}}$ \\ \hhline{=|=|=||=|=||=}
Softmax                   & 90.11           & 74.33          & 0.4110          & 0.6880              &  -  \\ \hhline{-|-|-||-|-||-}
Softmax+Center Loss       & 90.29           & 75.00          & \textbf{0.3295}          & 0.547           & 0.3893     \\ \hhline{-|-|-||-|-||-}
Softmax+Contrastive Center Loss  & \textbf{90.83}  & \textbf{76.63} & 0.4548    & \textbf{0.8127}  & \textbf{0.6122} \\ \hhline{-|-|-||-|-||-}
\end{tabular}
\caption{Compare Center Loss and the proposed CCL.}
\label{compare_cl_ccl}
\end{table}

\begin{table}[!h]
\centering
\begin{tabular}{C{1.4cm}|C{1.4cm}|C{1.4cm}|C{1.4cm}|C{1.4cm}|C{1.4cm}}
\hline
\textbf{alpha} & \textbf{beta} & \textbf{rank-1} & \textbf{mAP}   & $\boldsymbol{D_{intra}}$ & $\boldsymbol{D_{inter}}$ \\ \hline \hline
1                               & 1              & 83.49                            & 61.7                            & 0.4729 & 1.1591 \\ \hline
1                               & 0.1            & 90.2                             & 75.56                           & 0.2294 & 0.413  \\ \hline
0.1                             & 0.1            & \textbf{90.83}                   & \textbf{76.63}                  & 0.4548 & 0.8127 \\ \hline
0.1                             & 0.01           & 90.44                            & 75.36                           & 0.3806 & 0.6525 \\ \hline
0.01                            & 0.01           & 90.38                            & 76.07                           & 0.4168 & 0.7093 \\ \hline
\end{tabular}
\caption{Effect of different parameters on Market-1501.}
\label{ccl_parameters}
\end{table}

\subsection{Comparing with State-of-the-Art}
\label{final_compare}
According to Table~\ref{ccl_parameters}, we set $\alpha=0.1$ and $\beta=0.1$ in all following experiments.
We now compare our results with other state-of-the-art methods in Table~\ref{LARGE_RESULT} and ~\ref{SMALL_RESULT}.
With the single-query settings, our model achieves $90.8\%$ rank-1 accuracy and $76.6\%$ mAP on Market-1501.
On Duke-MTMC-reID, compared to the previous best model, we achieve an absolute improvement of $3.7\%$ in rank-1 and $6.7\%$ in mAP.
On CUHK03, the proposed model achieves $62.5\%$ rank-1 accuracy / $62.7\%$ mAP on CUHK03 (labeled), and $55.9\%$ rank-1 accuracy / $55.0\%$ mAP on CUHK03 (detected).
Another observation is that the performance on all datasets can be boosted by simply adopting a more powerful network, such as ResNet-101.
% These results demonstrate that the proposed deep antithetical learning does not require a specific network.
Therefore, it has a potential to serve as a practical method for boosting many existing ReID methods.

\begin{table}[h]
\centering
\begin{tabular}{c|cccc|cccc}
\hline
\multirow{2}{*}{\textbf{Methods}} & \multicolumn{4}{c|}{\textbf{Market-1501}}                              & \multicolumn{4}{c}{\textbf{Duke}}                                      \\ \cline{2-9} 
                         & \textbf{rank-1}  & \textbf{rank-5}  & \textbf{rank-10}  & \textbf{mAP}   & \textbf{rank-1}  & \textbf{rank-5}  & \textbf{rank-10}       & \textbf{mAP}           \\ \hline \hline
% MSCAN~\cite{MSCAN}                    & 80.3          & -             & -             & 57.5          & -             & -             & -             & -             \\ \hline
Re-rank~\cite{RERANK}                  & 77.1         & -             & -             & 63.6         & -             & -             & -             & -             \\ \hline
LSRO~\cite{LSRO}    & 84.0          & -             & -             & 66.1          & 67.7          & -             & -             & 47.1          \\ \hline
TriNet~\cite{TriNet}                   & 84.9          & 94.2          & -             & 69.1          & -             & -             & -             & -             \\ \hline
% TriNet+Re-rank           & 86.7          & 93.4          & -             & 81.1          & -             & -             & -             & -             \\ \hline
SVDNet~\cite{SVD}                   & 82.3          & 92.3          & 95.2          & 62.1          & 76.7          & 86.4          & 89.9          & 56.8          \\ \hline
DPFL~\cite{DPFL}                     & 88.6          & -             & -             & 72.6          & 79.2          & -             & -             & 60.6          \\ \hline \hline
Ours(ResNet50)      & 90.8  & \textbf{96.9}   & \textbf{98.0}   & 76.6        & 82.9         & 91.9         & 93.8   &67.3      \\ \hline
Ours+Rerank    & \textbf{92.7} & 95.8 & 97.2     & \textbf{89.1} & \textbf{87.4} & \textbf{92.5} & \textbf{94.6} & \textbf{83.2} \\\hline \hline
Ours(ResNet101)    & 91.5 & 96.8 & 97.7 & 79.5 & 84.5 & 92.6 & 94.3 & 70.1 \\ \hline
\end{tabular}
\caption{Results on Market1501 and Duke-MTMC-reID in single query mode.}
\label{LARGE_RESULT}
\end{table}

% \usepackage{multirow}
\begin{table}[h]
\centering
\caption{Results on CUHK03 (labeled) and CUHK03 (detected).}
\label{SMALL_RESULT}
\begin{tabular}{c|C{2.0cm}C{1.7cm}|C{1.7cm}C{1.7cm}}
\hline
\multirow{2}{*}{\textbf{Methods}} & \multicolumn{2}{c|}{\textbf{CUHK03 (labeled)}}  & \multicolumn{2}{c}{\textbf{CUHK03 (detected)}} \\ \cline{2-5} 
                         & \textbf{rank-1}  & \textbf{mAP}  & \textbf{rank-1}   & \textbf{mAP}           \\ \hline \hline
IDE+DaF~\cite{IDEDaF}                  & 27.5          & 31.5          & 26.4          & 30.0          \\ \hline
PAN~\cite{PAN}                      & 36.9          & 35.0          & 36.3          & 34.0          \\ \hline
DPFL~\cite{DPFL}                     & 43.0          & 40.5          & 40.7          & 37.0          \\ \hline
SVDNet~\cite{SVD}                   & 40.9          & 37.8          & 41.5          & 37.3          \\ \hline
TriNet~\cite{TriNet}                   & 58.1          & 53.8          & 55.5          & 50.7          \\ \hline
\hline
Ours(ResNet50)           & 62.5          & 62.7          & 55.9          & 55.0         \\ \hline
Ours+Rerank          & \textbf{68.3}         & \textbf{69.5}          & \textbf{59.6}          & \textbf{61.6}  \\ \hline \hline
Ours(ResNet101)           & 68.9          & 68.7          & 58.6          & 59.0         \\ \hline
\end{tabular}
\end{table}

\section{Conclusions}
In this paper, we analyze the ubiquitous image resolution discrepancy problem in person ReID tasks.
Extensive experiments indicate that these discrepancies have a negative impact on the ReID performance, and some mining strategies such as OHM will make this problem even worse.
In this paper, we propose a novel training framework called deep antithetical learning and address this issue in two steps.
First, an additional antithetical training set is generated for balancing biased resolution discrepancies. %  in the original training set.
Second, we propose a resolution-invariant objective function called Contrastive Center Loss. %  for producing better estimations of the image space.
% Experiment results demonstrate the effectiveness of the proposed framework relieving resolution discrepancies.
% Even using a vanilla ReID network, the proposed framework outperforms previous state-of-the-art methods by a large margin.
Experiments demonstrate that even using a vanilla ReID network, the proposed framework outperforms previous state-of-the-art methods by a large margin.


\begin{thebibliography}{8}

\bibitem{FPNN}
Li, W., Zhao, R., Xiao, T., Wang, X.:
\newblock Deepreid: Deep filter pairing neural network for person
  re-identification.
\newblock In: CVPR. (2014)

\bibitem{DLR}
Jiao, J., Zheng, W.S., Wu, A., Zhu, X., Gong, S.:
\newblock Deep low-resolution person re-identification.
\newblock In: AAAI. (2018)

\bibitem{IDLA}
Ahmed, E., Jones, M., Marks, T.K.:
\newblock An improved deep learning architecture for person re-identification.
\newblock In: CVPR. (2015)

\bibitem{CONTRASTIVELOSS}
Hadsell, R., Chopra, S., LeCun, Y.:
\newblock Dimensionality reduction by learning an invariant mapping.
\newblock In: CVPR. (2006)

\bibitem{ResNet}
He, K., Zhang, X., Ren, S., Sun, J.:
\newblock Deep residual learning for image recognition.
\newblock In: CVPR. (2016)

\bibitem{IDENTITY+VERIFY}
Zheng, Z., Zheng, L., Yang, Y.:
\newblock A discriminatively learned cnn embedding for person reidentification.
\newblock ACM TOMM (2017)

\bibitem{CENTERLOSS}
Wen, Y., Zhang, K., Li, Z., Qiao, Y.:
\newblock A discriminative feature learning approach for deep face recognition.
\newblock In: ECCV. (2016)

\bibitem{IDEDaF}
Yu, R., Zhou, Z., Bai, S., Bai, X.:
\newblock Divide and fuse: A re-ranking approach for person re-identification.
\newblock arXiv preprint arXiv:1708.04169 (2017)

\bibitem{ResNeXt}
Xie, S., Girshick, R., Doll{\'a}r, P., Tu, Z., He, K.:
\newblock Aggregated residual transformations for deep neural networks.
\newblock In: CVPR. (2017)

\bibitem{TriNet}
Hermans, A., Beyer, L., Leibe, B.:
\newblock In defense of the triplet loss for person re-identification.
\newblock arXiv preprint arXiv:1703.07737 (2017)

\bibitem{LSRO}
Zheng, Z., Zheng, L., Yang, Y.:
\newblock Unlabeled samples generated by gan improve the person
  re-identification baseline in vitro.
\newblock arXiv preprint arXiv:1701.07717 \textbf{3} (2017)

\bibitem{CUHK03}
Li, W., Zhao, R., Xiao, T., Wang, X.:
\newblock Deepreid: Deep filter pairing neural network for person
  re-identification.
\newblock In: CVPR. (2014)

\bibitem{DUKEMTMCREID}
Zheng, Z., Zheng, L., Yang, Y.:
\newblock Unlabeled samples generated by gan improve the person
  re-identification baseline in vitro.
\newblock In: ICCV. (2017)

\bibitem{MARKET}
Zheng, L., Shen, L., Tian, L., Wang, S., Wang, J., Tian, Q.:
\newblock Scalable person re-identification: A benchmark.
\newblock In: ICCV. (2015)

\bibitem{PAN}
Zheng, Z., Zheng, L., Yang, Y.:
\newblock Pedestrian alignment network for large-scale person
  re-identification.
\newblock arXiv preprint arXiv:1707.00408 (2017)

\bibitem{DPFL}
Chen, Y., Zhu, X., Gong, S.:
\newblock Person re-identification by deep learning multi-scale
  representations.
\newblock In: CVPR. (2017)

\bibitem{LOMO}
Liao, S., Hu, Y., Zhu, X., Li, S.Z.:
\newblock Person re-identification by local maximal occurrence representation
  and metric learning.
\newblock In: CVPR. (2015)

\bibitem{SVD}
Sun, Y., Zheng, L., Deng, W., Wang, S.:
\newblock Svdnet for pedestrian retrieval.
\newblock In: CVPR. (2017)

\bibitem{LBP}
Ojala, T., Pietikainen, M., Maenpaa, T.:
\newblock Multiresolution gray-scale and rotation invariant texture
  classification with local binary patterns.
\newblock PAMI \textbf{24} (2002)  971--987

\bibitem{GAN}
Goodfellow, I., Pouget-Abadie, J., Mirza, M., Xu, B., Warde-Farley, D., Ozair,
  S., Courville, A., Bengio, Y.:
\newblock Generative adversarial nets.
\newblock In: NIPS. (2014)

\bibitem{DCGAN}
Radford, A., Metz, L., Chintala, S.:
\newblock Unsupervised representation learning with deep convolutional
  generative adversarial networks.
\newblock arXiv preprint arXiv:1511.06434 (2015)

\bibitem{COLORHIS1}
Farenzena, M., Bazzani, L., Perina, A., Murino, V., Cristani, M.:
\newblock Person re-identification by symmetry-driven accumulation of local
  features.
\newblock In: CVPR. (2010)

\bibitem{MSCAN}
Li, D., Chen, X., Zhang, Z., Huang, K.:
\newblock Learning deep context-aware features over body and latent parts for
  person re-identification.
\newblock In: CVPR. (2017)

\bibitem{SGD}
Bottou, L.:
\newblock Large-scale machine learning with stochastic gradient descent.
\newblock In: COMPSTAT.
\newblock (2010)

\bibitem{KISSME}
Koestinger, M., Hirzer, M., Wohlhart, P., Roth, P.M., Bischof, H.:
\newblock Large scale metric learning from equivalence constraints.
\newblock In: CVPR. (2012)

\bibitem{RANDOMERASE}
Zhong, Z., Zheng, L., Kang, G., Li, S., Yang, Y.:
\newblock Random erasing data augmentation.
\newblock arXiv preprint arXiv:1708.04896 (2017)

\bibitem{DML}
Yi, D., Lei, Z., Liao, S., Li, S.Z.:
\newblock Deep metric learning for person re-identification.
\newblock In: ICPR. (2014)

\bibitem{RERANK}
Zhong, Z., Zheng, L., Cao, D., Li, S.:
\newblock Re-ranking person re-identification with k-reciprocal encoding.
\newblock In: CVPR. (2017)

\bibitem{ITML}
Davis, J.V., Kulis, B., Jain, P., Sra, S., Dhillon, I.S.:
\newblock Information-theoretic metric learning.
\newblock In: ICML. (2007)

\bibitem{MEASURESHARP}
De, K., Masilamani, V.:
\newblock Image sharpness measure for blurred images in frequency domain.
\newblock Procedia Engineering \textbf{64} (2013)  149--158

\bibitem{COLORHIS2}
Zhao, R., Ouyang, W., Wang, X.:
\newblock Unsupervised salience learning for person re-identification.
\newblock In: CVPR. (2013)

\bibitem{COLORNAMES1}
Yang, Y., Yang, J., Yan, J., Liao, S., Yi, D., Li, S.Z.:
\newblock Salient color names for person re-identification.
\newblock In: ECCV. (2014)

\bibitem{NPSM}
Liu, H., Feng, J., Jie, Z., Jayashree, K., Zhao, B., Qi, M., Jiang, J., Yan,
  S.:
\newblock Neural person search machines.
\newblock In: ICCV. (2017)

\bibitem{DeepAligned}
Zhao, L., Li, X., Zhuang, Y., Wang, J.:
\newblock Deeply-learned part-aligned representations for person
  re-identification.
\newblock In: CVPR. (2017)

\bibitem{NR-IQA}
De, K., Masilamani, V.:
\newblock Image sharpness measure for blurred images in frequency domain.
\newblock Procedia Engineering \textbf{64} (2013)  149--158

\bibitem{SRGAN}
Ledig, C., Theis, L., Huszar, F., Caballero, J., Cunningham, A., Acosta, A.,
  Aitken, A., Tejani, A., Totz, J., Wang, Z.,  et~al.:
\newblock Photo-realistic single image super-resolution using a generative
  adversarial network.
\newblock In: CVPR. (2017)

\bibitem{DeblurGAN}
Kupyn, O., Budzan, V., Mykhailych, M., Mishkin, D., Matas, J.:
\newblock Deblurgan: Blind motion deblurring using conditional adversarial
  networks.
\newblock arXiv preprint arXiv:1711.07064 (2017)

\bibitem{BeyondPartModel}
Sun, Y., Zheng, L., Yang, Y., Tian, Q., Wang, S.:
\newblock Beyond part models: Person retrieval with refined part pooling.
\newblock arXiv preprint arXiv:1711.09349 (2017)

\bibitem{AlignedReID}
Zhang, X., Luo, H., Fan, X., Xiang, W., Sun, Y., Xiao, Q., Jiang, W., Zhang,
  C., Sun, J.:
\newblock Alignedreid: Surpassing human-level performance in person
  re-identification.
\newblock arXiv preprint arXiv:1711.08184 (2017)

\bibitem{CycleGAN}
Zhu, J.Y., Park, T., Isola, P., Efros, A.A.:
\newblock Unpaired image-to-image translation using cycle-consistent
  adversarial networks.
\newblock In: CVPR. (2017)

\bibitem{MSML}
Xiao, Q., Luo, H., Zhang, C.:
\newblock Margin sample mining loss: A deep learning based method for person
  re-identification.
\newblock arXiv preprint arXiv:1710.00478 (2017)

\bibitem{EdgeDetection}
Marziliano, P., Dufaux, F., Winkler, S., Ebrahimi, T.:
\newblock Perceptual blur and ringing metrics: application to jpeg2000.
\newblock Signal processing: Image communication \textbf{19} (2004)  163--172

\bibitem{HVS}
Du, J., Yu, Y., Xie, S.:
\newblock A new image quality assessment based on hvs.
\newblock Journal of Electronics (China) \textbf{22} (2005)  315--320

\bibitem{WAVELET1}
Sheikh, H.R., Bovik, A.C., Cormack, L.:
\newblock No-reference quality assessment using natural scene statistics:
  Jpeg2000.
\newblock TIP \textbf{14} (2005)  1918--1927

\bibitem{WAVELET2}
Chen, M.J., Bovik, A.C.:
\newblock No-reference image blur assessment using multiscale gradient.
\newblock EURASIP Journal on image and video processing \textbf{2011} (2011) ~3

\bibitem{NSS}
Brand{\~a}o, T., Queluz, M.P.:
\newblock No-reference image quality assessment based on dct domain statistics.
\newblock Signal Processing \textbf{88} (2008)  822--833

\bibitem{InDefense}
Hermans, A., Beyer, L., Leibe, B.:
\newblock In defense of the triplet loss for person re-identification.
\newblock arXiv preprint arXiv:1703.07737 (2017)

\end{thebibliography}
\end{document}